\documentclass{article}

\usepackage{xcolor}
\usepackage{amsmath} 
\usepackage{graphicx}
\graphicspath{ {./} }%{ {../plots/} }
\usepackage{subcaption}
 \usepackage[nonatbib,final]{investment_vs_reward}

\usepackage[utf8]{inputenc} % allow utf-8 input
\usepackage[T1]{fontenc}    % use 8-bit T1 fonts
\usepackage{hyperref}       % hyperlinks
\usepackage{url}            % simple URL typesetting
\usepackage{booktabs}       % professional-quality tables
\usepackage{amsfonts}       % blackboard math symbols
\usepackage{nicefrac}       % compact symbols for 1/2, etc.
\usepackage{microtype}      % microtypography

\title{Investment vs. reward in a competitive knapsack problem}

\author{%
  Oren Neumann \\
  Institute for Theoretical Physics\\
  Goethe University Frankfurt\\
  Frankfurt am Main, Germany \\
  \texttt{neumann@itp.uni-frankfurt.de} \\
  \And
  Claudius Gros \\
  Institute for Theoretical Physics\\
  Goethe University Frankfurt\\
  Frankfurt am Main, Germany \\
  \texttt{gros@itp.uni-frankfurt.de} \\
}

\begin{document}

\maketitle

\begin{abstract}
Natural selection drives species to develop brains,
with sizes that increase with the complexity of the 
tasks to be tackled. Our goal is to investigate
the balance between the metabolic costs of larger brains
compared to the advantage they provide in solving 
general and combinatorial problems. Defining advantage 
as the performance relative to competitors, a two-player 
game based on the knapsack problem is used. Within this
framework, two opponents compete over shared resources, 
with the goal of collecting more resources than the 
opponent. Neural nets of varying sizes are trained 
using a variant of the AlphaGo Zero algorithm 
\cite{silver2017mastering}. A surprisingly simple relation, 
$N_A/(N_A+N_B)$, is found for the relative win rate 
of a net with $N_A$ neurons against one with 
$N_B$. Success increases linearly with investments 
in additional resources when the networks sizes 
are very different, i.e.\ when $N_A \ll N_B$, with 
returns diminishing when both networks become comparable 
in size. 
\end{abstract}

%------------------------------------------------
\section{Introduction}
%------------------------------------------------
Optimal resource allocation often leads to combinatorial 
problems, e.g.\ time can be minimized by solving
the travelling salesman problem. In nature, resource
allocation is in many cases a competitive process, with 
animals competing with each other over shared resources. A key question regards 
in this context the balance between success rate and
investments in computational capabilities, e.g.\ in terms 
of metabolic cost. This question is relevant also for
artificial neural networks, which are notoriously bad 
at solving combinatorial problems \cite{smith1999neural}, in 
particular when compared with the performance of 
traditional, problem-specific algorithms.

Lately, there has been a rising interest to apply
deep learning architectures to combinatorial problems 
\cite{abe2019solving} \cite{khalil2017learning} \cite{bello2016neural}. In this context
we are interested in the evolutionary factors determining 
network sizes, when assuming that larger nets come with
correspondingly larger metabolic cost. Presently it
is however unclear how costs scale with reward, that
is the precise functional dependence of success on network
size is not known. Here we investigate this question
within the 0-1 knapsack problem, a basic resource 
allocation task that can be generalize to a range 
of real-world situations \cite{murawski2016humans}. 

The problem is defined as follows:
Given $n$ items with values $v_i$ and weights $w_i$, maximize
\begin{equation}
\sum_{i=1}^{n}{v_i x_i}, \quad\quad x_i \in \{0,1\}\,,
\end{equation}
subject to a capacity constraint
\begin{equation}
\sum_{i=1}^{n}{w_i x_i} \,\leq\, W\,,
\end{equation}
where $\{x_i\}$ denotes the chosen subset of items and 
$W$ is a constraint on the total weight. In other words, 
one should select the group of items with maximal total 
value which does not exceed the total weight limit. There 
are several algorithms developed specifically to solve 
the 0-1 knapsack problem and special cases of it, such 
as dynamic programming and branch-and-bound algorithms \cite{martello2000new}.

Survival of the fittest pushes species to become better 
than their competition, when too many individuals are 
fighting for the same resources. It is therefore logical
to measure the success of an agent only in comparison to 
other agents, rather than looking for an absolute scale. 
For this reason we decided to focus on a two-player 
extension of the knapsack problem. In this setup, two 
agents compete over the same pool of resources trying each
to collect more value than the other. This framework 
allowed us to use a powerful tool for training deep 
learning models to play turn based games, the AlphaGo Zero 
($\alpha$GZ) algorithm \cite{silver2017mastering}. Suitably adapted, 
we applied $\alpha$GZ to train neural network models of varying 
sizes $N$ to play the two-player knapsack game, with the goal 
of comparing their performances when playing against each 
other. This framework allows us to determine how increasing 
the number of available neurons affects the
performance of the competing networks.

%------------------------------------------------
\section{Method}
%------------------------------------------------
In order to simulate direct interactions between agents, we created a zero sum
game for two players which is an extension of the knapsack problem. During the
course of a game, players pick items in turns from a common item pool. Each
item can be selected by at most one player, and each player has a capacity
limit for the total weight of items they can personally collect. The goal of
the game is not to maximize the total value of items collected, but rather to
surpass the total value gained by the opponent.  A game progresses in the
following manner:
\begin{itemize}
\item Each turn, a player may choose any item from the pool of free items and
add it to their collection, provided this would not cause the total weight of
items collected to exceed the weight limit. The item chosen is then removed
from the common pool, and the other player plays his turn. If no suitable item
is available, the current player passes.  

\item The two players take their turns
one after the other, until no item is picked for two consecutive turns, meaning
there are no more valid items. At this point the player with
the larger total value is declared the winner.
\end{itemize}

For our simulations we focused on games where items are unique, 
$v_i\ne v_j$ for $i\ne j$, with both players having identical 
capacity limits. Since generating $v_i , w_i$ from a uniform distribution creates mostly easily solvable knapsack problem instances  \cite{smith2012measuring}, we generated weakly correlated instances \cite{pisinger2005hard}, which are generally harder to solve. These are defined as instances where $w_i$ is uniformly  distributed in $[0,1]$ and $v_i$ is uniformly distributed in a range of $w_i \pm 0.1$, confined to $[0,1]$.
The capacity limits were set to $W=n/4$ as this avoids generating easy instances in the normal knapsack problem setting \cite{ohlsson1993neural}.
The complexity of the
two player framework is somewhat increased, as compared to the 
original knapsack problem, as players need to deprive the
opponent of items, in addition to optimizing their own collection.

Training was done using an adaptation of the $\alpha$GZ
algorithm. The neural network agents are trained by 
reinforcement learning on self generated data, without using 
any a priori knowledge of the game. The networks receive as 
inputs the current state of the game, i.e.\ the weights and values 
of all items and a list of all items acquired so far by each
player. The output consists of a policy vector and a value prediction:
\begin{equation}
f(s) = (\vec{p},v)\,,
\end{equation}
where $v$ is a prediction of the value of the current game position $s$, which
is equivalent to the expected final outcome of the game from this position. We
set the possible game outcomes to be $\{0,1\}$, for losing/winning the game
respectively, therefore implying that $v \in [0,1]$. $\vec{p}$ is a probability
distribution over all available moves, which should prioritize better moves and
is used to guide the Monte Carlo tree search (MCTS). 
The neural net is trained on randomly picked game states that were 
visited in previous games played, using the loss function:
\begin{equation}
l = (z-v)^2 - \vec{\pi}\log\vec{p}\,.
\end{equation}
Here
$z$ is the recorded game outcome and $\vec{\pi}$ is an improved policy
vector generated by the MCTS.  Before each training phase, several games are
played using the current version of the agent, pitched against a copy of
itself. During a game, each player makes use of a search tree in order to
choose every move. The tree contains a parent node holding information about
the current game state, with nodes descending from it for each  possible future
game state that has been explored before. Every turn, the agent is given a
certain number of iterations to expand their search tree, before deciding which
move to make. 

MCTS makes use of the information stored in each node in order to expand new
leaf nodes. A node contains four parameters: The number of times the node has
been visited in previous searches, $N$; the sum of all scores predicted for
this node and its children, $W$; the mean score $Q={W}/{N}$; and the policy
vector $\vec{p}$ calculated once by the agent for this game state.  Each
iteration of expanding the tree traverses it by starting at the first node,
i.e.\ the current game state, and repeatedly moving to the child node which
maximizes a quantity $Q+U$ until a previously unvisited node is reached. The
term $U$ is defined according to:
\begin{equation}
U = c P_a\frac{\sqrt{\sum_i N_i}}{1+N_a}\,,
\end{equation}
where $c$ is a constant, $P_a$ is the policy vector element 
corresponding to action $a$ leading to the node, $N_a$ is 
the node's visit count and $\sum_i N_i$ is the sum of the 
visit counts of all available actions.  When an unexplored
node is reached, it is added to the tree and the parameters 
of it and all other nodes traversed are updated. After the 
expansion iterations are done a move is chosen according 
to an improved policy $\vec{\pi}$, defined by using the visit
counts of all possible moves:
\begin{equation}
\pi_i \propto N_i^{1/\tau}
\end{equation}
With $\tau$ a temperature parameter controlling the degree of exploration, set
to $1$ in our case. The turn then ends, and both players trim their search
trees to remove all nodes that are now unreachable, making the new game state
the parent node of the whole tree.

We trained fully connected feed forward neural network models to play two-player 
knapsack games with 16 items. Each net had 2 hidden layers with ReLU connections, both with the 
same number of neurons, which changed from one net to the other.
The inputs consisted of 4 vectors of length 16: A vector of the items' weights, 
a vector of the values and two binary vectors encoding the items taken by each player.
The items were ordered by descending ratio of $v_i/w_i$ such that a minimal sized 
model could easily implement a greedy strategy of picking the items with the best ratios.
The network outputs were a policy vector with a softmax activation and a value 
head with a sigmoid activation.
Every net was trained on  a total of 40,000 
self generated games, where optimization steps were taken every 
40 games and performance was evaluated every 4,000 games, saving 
the best performing version. Each evaluation step consisted of 
200 games of the net against a greedy algorithm in order to save 
time. The agents used 40 Monte Carlo steps per turn in all games. 
All model instances were given the same number of optimization 
steps regardless of their sizes, without early stopping if performance saturated.
The final results presented in the next section were generated 
by training four separate copies of each net and taking the average 
of their game outcomes.

In order to interpret our results we made use of the Elo rating system
\cite{glickman1999rating}, a rating system for zero-sum games that was invented  for
chess and gained popularity in numerous games. The expected score, or winning
probability, of player $A$ over player $B$ is given by the formula:
\begin{equation} \label{elo}
P_A = \frac{1}{1 + 10^{(R_B - R_A)/400}}\,,
\end{equation}
where $R_A, R_B$ are the Elo ratings of both players. These ratings are found
by repeatedly matching players together and adjusting their ratings to fit the
game outcomes.

%------------------------------------------------
\section{Results and discussion}
%------------------------------------------------

%---------------------------------------
\begin{figure}[t]
	\begin{subfigure}{.5\textwidth}
  		\centering
  		\includegraphics[width=1.1\linewidth]{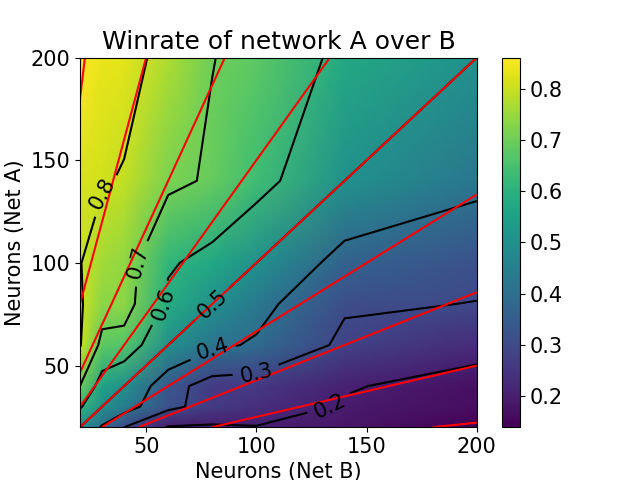}
  		\caption{} 
  		\label{fig:sfig1}
	\end{subfigure}%
	\begin{subfigure}{.5\textwidth}
  		\centering
  		\includegraphics[width=1.1\linewidth]{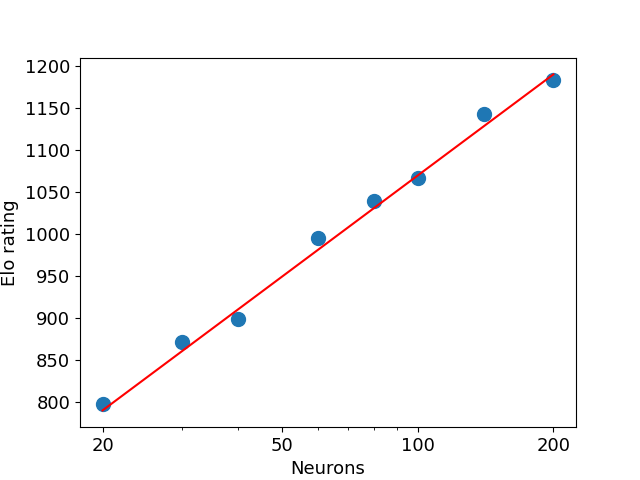}
  		\caption{}
  		\label{fig:sfig2}
	\end{subfigure}
\caption{Training neural networks to play competitive knapsack problem games
with 16 items. (a) Probability of net A to win against net B, when both
have differing numbers of neurons. The black lines mark increments of 0.1 in
winning probability, the red lines mark the same for the prediction given by
Eq.~( \ref{win_prob}). (b) Elo rating of networks 
with different numbers of neurons in log scale, with the logarithmic 
fit $y=400\log(N)+const$ in red.}
\label{fig:16_competition}
\end{figure}
%---------------------------------------

Combinatorial problems manifest themselves in many 
ways in nature, making it beneficial for organisms 
to evolve ways to solve them. But unlike raw
mathematical problems which have in most cases goals 
and solving conditions that are defined in absolute terms, 
in the natural world evolution pushes agents to overcome 
their competition such that merit is measured predominantly 
in relation to one's competitors. It is therefore biologically 
sensible to look at adversarial problems when
analysing the performance of neural networks at solving 
combinatorial problems. This is why we chose to work on 
the two-player competitive version of the knapsack problem, 
where the merit of an agent in relation to other agents is
clearly defined by the probability of winning or losing a 
game against them. Figure \ref{fig:16_competition} presents 
the outcome of training neural nets of varying sizes to play 
a competitive knapsack problem game with 16 weakly correlated 
items \cite{pisinger2005hard}. All neural nets contained two 
hidden layers, both with the same number of neurons, which 
was changed from one net to the other.

The problem displays diminishing returns, requiring increasingly larger
differences between the sizes of competing nets in order to maintain the same
win-lose statistics between them as they increase in size. This is supported by
plotting the Elo rating of the nets, which scales logarithmically in network size.
In fact the Elo rating is fitted well by
$y=400\log(N)+const$, where $N$ is the number of hidden neurons. Since $400$ is 
the Elo scaling factor, plugging this expression into 
Eq.~(\ref{elo}) reveals that the probability of net A to win a 
game against net B reduces to 
\begin{equation} \label{win_prob}
P_A = \frac{N_A}{N_A+N_B}\,,
\end{equation}
where $N_A, N_B$ are the numbers of hidden neurons of nets A 
and B respectively. In other words, the 
probability of player A to win is $N_A / N_B$ times greater 
than that of player B, meaning the performance of neural nets
against each other in this task is determined directly by 
the ratio of their sizes.

A peculiar property of the Elo rating system is that it 
assumes the existence of an absolute rating scale which 
determines game outcomes solely according to the difference 
in ratings between the two players. It could be possible that 
other factors also affect game outcomes, which would require 
a more complex model. To rule out this possibility, we plotted 
the full map of game outcomes for every combination of players 
in Figure \ref{fig:sfig1}. Lines of equal winning probability 
predicted by Eq.~(\ref{win_prob}) are marked in red. It is clear 
that the theoretical prediction matches the corresponding lines
obtained through simulations (in black) to an astonishing degree, 
proving that Eq.~(\ref{win_prob}) holds.

This result implies a linear increase of the probability $P_A$ with network size in the regime
where $N_A \ll N_B$, and diminishing returns when both opponents are relatively
equally matched. Therefore a large difference between competitors can be
quickly closed when they evolve to maximize their own utility.  Note that when
$N_A \approx N_B$ the benefit of increasing the number of neurons by a
multiplicative factor, $N_A \to \gamma N_A$, is independent of $N_A$, always
yielding the same winning probability ${\gamma}/(\gamma+1)$.

%%%%%%%%%%%%%%%%%%%%%%%%%%%%%%%%%%%%%%%%%%%%%

%\bibliography{references}{}
%\bibliographystyle{unsrt}

\end{document}